%% file: main.tex
\documentclass[sigconf,10pt,anonymous=false]{acmart}

\AtBeginDocument{%
  }

\setcopyright{acmlicensed}
\copyrightyear{2024}
\acmYear{2024}
\acmDOI{XXXXXXX.XXXXXXX}

\acmConference[XXX '25]{Make sure to enter the correct
  conference title from your rights confirmation emai}{June 03--05,
  2018}{Woodstock, NY}

\usepackage{enumitem}
\usepackage{multirow} 
\usepackage{array}
\usepackage{multirow}
\usepackage{graphicx} 
\usepackage{makecell}
\usepackage{subcaption}
\usepackage{comment}
\usepackage{bm}

\usepackage{titlesec}

\titlespacing{\section}{0pt}{4 pt}{2 pt}
\titlespacing{\subsection}{0pt}{4 pt}{2 pt}
\titlespacing{\subsubsection}{0pt}{3 pt}{1 pt}

\acmISBN{978-1-4503-XXXX-X/18/06}

\begin{document}

\title{Foundation Models for CPS-IoT:\\Opportunities and Challenges}

\author{Ozan Baris$^1$, Yizhuo Chen$^2$, Gaofeng Dong$^3$, Liying Han$^3$,\\Tomoyoshi Kimura$^2$, Pengrui Quan$^3$, Ruijie Wang$^2$, Tianchen Wang$^2$,\\Tarek Abdelzaher$^2$, Mario Bergés$^1$, Paul Pu Liang$^4$, Mani Srivastava$^3$}
\affiliation{
  \vspace{2mm}
  \institution{$^1$CMU, $^2$UIUC, $^3$UCLA, $^4$MIT}
  \country{}
}

\authornote{All authors contributed equally to this research. Junior researchers are listed first in the alphabetical order of their last names, and subsequently the senior researchers similarly. Bergés and Srivastava hold concurrent appointments as Amazon Scholars, and as Professors at their respective universities, but the work in this paper is not associated with Amazon. Contact Author: Mani Srivastava <mbs@ucla.edu>.}

\renewcommand{\shortauthors}{Baris et al.}

\renewcommand{\shorttitle}{Foundation Models for CPS-IoT: Opportunities and Challenges}

\def\mbs#1{{\color{red}#1}} 
\def\mbsdel#1{} 
\def\tkimuradel#1{} 

\begin{abstract}

Methods from machine learning (ML) have transformed the implementation of Perception-Cognition-Communication-Action loops in Cyber-Physical Systems (CPS) and the Internet of Things (IoT), replacing mechanistic and basic statistical models with those derived from data. However, the first generation of ML approaches, which depend on supervised learning with annotated data to create task-specific models, faces significant limitations in scaling to the diverse sensor modalities, deployment configurations, application tasks, and operating dynamics characterizing real-world CPS-IoT systems. The success of task-agnostic foundation models (FMs), including multimodal large language models (LLMs), in addressing similar challenges across natural language, computer vision, and human speech has generated considerable enthusiasm for and exploration of FMs and LLMs as flexible building blocks in CPS-IoT analytics pipelines, promising to reduce the need for costly task-specific engineering. 


Nonetheless, a significant gap persists between the current capabilities of FMs and LLMs in the CPS-IoT domain and the requirements they must meet to be viable for CPS-IoT applications. In this paper, we analyze and characterize this gap through a thorough examination of the state of the art and our research, which extends beyond it in various dimensions. Based on the results of our analysis and research, we identify essential desiderata that CPS-IoT domain-specific FMs and LLMs must satisfy to bridge this gap. We also propose actions by CPS-IoT researchers to collaborate in developing key community resources necessary for establishing FMs and LLMs as foundational tools for the next generation of CPS-IoT systems.
\end{abstract}




\maketitle

\input{sections/1-Introduction}

\input{sections/2-SOTA}

\input{sections/3-Experience}
\input{sections/4-QuoVadis}
\input{sections/5-Conclusions}


\begin{acks}
This research was funded in part by the Air Force Office of Scientific Research under Cooperative Agreement \#~FA95502210193, the DEVCOM ARL under Cooperative Agreement \#~W911NF-17-2-0196, the NIH mDOT Center under Award \#~1P41EB028242, the NSF under Award \#~CNS-2325956 and \#~CNS-2038817, and the Boeing Company. It was
also supported in part by the Pennsylvania Infrastructure Technology Alliance (PITA) and by ACE, one of the seven centers
in JUMP 2.0, a Semiconductor Research Corporation (SRC)
program sponsored by DARPA. 
\end{acks}

\bibliographystyle{ACM-Reference-Format}
\bibliography{final_references}

\end{document}

%% file: sections/1-Introduction.tex
\vspace{-2mm}
\section{Introduction}


\noindent
Cyber-Physical Systems and the Internet-of-Things (referred to as CPS-IoT Systems henceforth) operate through \textit{Perception-Cognition-Communication-Action (PCCA)} loops, leveraging multimodal and multiview sensor data to comprehend physical states, predict future and remote conditions, exchange information with other systems, and execute timely interventions. Recently, CPS-IoT engineering has shifted from mechanistic and statistical models rooted in human expertise to high-dimensional, data-driven models.

Advances in machine learning (ML) and artificial intelligence (AI) research in language, vision, and speech have led to specialized neural models and training algorithms tailored for the CPS-IoT domain. These models demonstrate notable performance gains across common signal modalities like IMU, radar, lidar, and wireless, addressing tasks such as human activity recognition, acoustic event detection, spatio-temporal tracking, and signal generation. They leverage unique characteristics like spectral domain information while tackling challenges such as sensor placement sensitivity and varying sampling rates.

However, the first generation of ML-based CPS-IoT systems faces several critical limitations. First, the trained models are task-specific, designed for particular tasks and requiring dedicated retraining for each (new) task the system must perform. Second, this task-specific supervised training strategy requires copious amounts of labeled sensor data, which is difficult to obtain because many sensing modalities are not human-interpretable, making retrospective labeling challenging. Moreover, contemporaneous labeling is intrusive, privacy-invading, and low-quality, particularly as many CPS-IoT systems operate in real-time out in the wild; and due to the limited scope of deployments/studies, most labeling strategies typically lead to data imbalance issues. Lastly, these models are often dependent on specific sensor sampling rates, placements, and configurations that vary between deployments.

\textbf{Emergence of Foundation Models:} 
In broader AI/ML research and industry, there is a clear trend toward homogenizing model functionalities. Supervised discriminative models, limited by the high cost and scarcity of labeled data, are increasingly being replaced by generative models trained through self-supervised learning on generic pretext tasks. This shift has led to the rise of \textit{Foundation Models} (FMs), introduced in~\cite{bommasani2021opportunities}, which are defined as models trained on broad datasets using large-scale self-supervision and adaptable to a wide range of downstream tasks. FMs, such as Large Language Models (LLMs), have revolutionized AI system design by enabling task adaptation via FM-based pipelines rather than bespoke task-specific ones. The success of FMs is driven by three key factors: self-supervised pretraining, scale (of both data and computation), and innovative architectures like transformers~\cite{vaswani2017attention}, structured state-space models~\cite{gu2023mamba, dao2024transformers}, and hybrid approaches~\cite{lieber2024jamba}. While early FMs like BERT emphasized unifying representations requiring task-specific stages, later models with natural language interfaces, such as LLMs and Large Vision Language models, can specify and perform semantically unrelated tasks. Methods for adapting FMs to downstream tasks—ranging from full fine-tuning and task-specific adapters to input prompt tuning, in-context learning, and zero-shot prompting—have further broadened their applicability.

\textbf{Foundation Models for CPS-IoT Domain:} 
Mirroring the rise of FMs in NLP, vision, and speech, the CPS-IoT domain has seen parallel advancements in developing FMs tailored to its unique challenges. Indeed, the factors driving the emergence of FMs in broader AI/ML are even more critical in the CPS-IoT domain: sensor data types are more diverse, labeling is harder, unlabeled sensor data is far more abundant both absolutely and relatively, tasks are more varied with stricter performance and safety requirements, systems scale more extremely, and platforms are more resource-constrained. Many have recognized the potential of FMs for CPS-IoT, spurring significant efforts to develop task-independent sensor time series representations (e.g.,
\texttt{LIMU-\\BERT}~\cite{xu2021limu}, \texttt{ImageBind}~\cite{girdhar2023imagebind}, \texttt{FOCAL}~\cite{liu2023focal}), leverage pretrained LLMs with their world knowledge, sequence processing, and spatiotemporal reasoning for general sensor analysis tasks (e.g., \texttt{Penetrative AI}~\cite{xu2024penetrative}, \texttt{IoT-LM}~\cite{mo2024iot}, \texttt{LLMSense}~\cite{ouyang2024llmsense}), and address specific tasks (e.g., \texttt{Chronos}~\cite{ansari2024chronos}, \texttt{MOMENT}~\cite{goswami2024moment} for forecasting) and applications (e.g., \texttt{RT-2}~\cite{brohan2023rt} for robot control and \texttt{LSM}~\cite{narayanswamy2024scaling} for wearable health tracking).

\textbf{About the Paper:} 
While there is considerable activity on CPS-IoT-related FMs and the use of LLMs for CPS-IoT, much of it directly extends models, architectures, training methods, and applications from general AI/ML domains like language and vision. At the same time, CPS-IoT systems have distinctive characteristics and needs -- arising from their embodied and embedded nature, and their tight coupling with the physical world -- which FMs must address. Unlike language, sensor data are discretized samples of continuous spatiotemporal physical signals rather than ordered symbolic sequences. Factors such as sampling rate, policy, quantization strategy, gain factor, location, and orientation affect both the quality and content of the physical world information in the sensor data stream. Likewise, CPS-IoT tasks must consider the spatiotemporal properties of sensor measurements and output actions, as well as latency in producing outputs.


Motivated by these observations and guided by an in-depth analysis of state-of-the-art work and insights from our preliminary research, this paper explores the desiderata for CPS-IoT FMs and lays out a research agenda. This agenda identifies key technical challenges to address and calls for collaborative efforts to create the necessary artifacts to realize it. The ideas in this paper draw on the collective expertise of the multi-institutional authoring team, spanning the entire CPS-IoT system stack and multiple applications, including prior contributions to FMs and LLMs for CPS-IoT.

The paper is divided into three main sections. Section~\ref{sec:sota} analyzes the state of the art in CPS-IoT-focused FMs and the use of LLMs for this purpose. Section~\ref{sec:our_research} presents preliminary findings from our research, addressing challenges such as resource-constrained platforms, sensor viewpoints, and tasks beyond prediction. Finally, Section~\ref{sec:quo_vadis} synthesizes insights from the earlier sections into a vision for future research and community-driven artifacts.

%% file: sections/2-SOTA.tex
\section{State of the Art in CPS-IoT FMs}
\label{sec:sota}
\noindent
Existing research in FMs (including LLMs) relevant to CPS-IoT has primarily focused on \textit{perception}, with limited attention to \textit{cognition, communication, and action} of the PCCA loops.
To assess the state-of-the-art (SOTA), we first examine perception-related FMs along two\tkimuradel{ key} axes: sensor modalities handled (single, multiple, and flexible) and tasks performed (fixed, configurable, selectable, and run-time specifiable). 
We then explore FMs beyond the perception tasks for CPS-IoT.


\subsection{CPS-IoT FMs for Perception: A Sensor Modality Perspective}
\noindent
Much of the research on CPS-IoT FMs for perception focuses on projecting raw sensor signals into compact representations, or \textit{embeddings}, that encode underlying physical phenomena and are generalizable to various downstream tasks.
Self-supervised learning (SSL) is commonly used to learn\tkimuradel{ task-independent} such representations without requiring task-specific annotations. The trained FM is then adapted to compute desired outputs for a specific downstream task, such as event classification or spatiotemporal localization. This stage is either learned from limited annotated data or designed from first principles and human knowledge.
However, learning robust representations for CPS-IoT systems is challenging due to the spatiotemporal characteristics of \tkimuradel{time-series} sensor signals~\cite{jin2023large, liu2023unsupervised} and the significant heterogeneity of sensor modalities across application domains~\cite{liang2024foundations}. 
We categorize prior works on SSL for CPS-IoT FMs into three groups, focusing on single\tkimuradel{ sensor} modality, a fixed set of \tkimuradel{multiple sensor} modalities, or a flexible set of\tkimuradel{ multiple sensor} modalities.



\subsubsection{Unimodal Models.}~Most CPS-IoT FMs focus on mapping a single sensor modality (e.g., image, IMU, sound) into embeddings.
SSL frameworks for these FMs either contrast temporally close samples to capture temporal consistency~\cite{yue2022ts2vec, tonekaboni2021unsupervised} or reconstruct masked portions of time-series~\cite{dong2023simmtm, nie2022time, xu2021limu}.
For human activity recognition\tkimuradel{, with wearable accelerometer sensor data}, the FM in~\cite{yuan2024self} uses a multi-task self-supervision approach employing the arrow of time, permutation, and time warping to enhance\tkimuradel{ downstream} generalization. 
Similarly, \texttt{RelCon}~\cite{xu2024relcon} applies a learnable distance measure and softened contrastive loss to capture motif similarity and domain-specific semantics\tkimuradel{ like rotation invariance}.
Alternatively, many SSL approaches explore frequency representations of raw signals to capture spatiotemporal features.
\texttt{TimesNet}~\cite{wu2022timesnet} transforms 1D time series into\tkimuradel{ a set of} 2D tensors\tkimuradel{ based on multiple periods} to capture multi-periodicity.
\texttt{TFC}~\cite{zhang2022self} enforces time-frequency consistency through contrastive learning. \texttt{AudioMAE}~\cite{huang2022masked} builds on MAE~\cite{he2022mae} architectures to reconstruct masked spectrograms of acoustic signals. \texttt{PhyMask}~\cite{PhyMask} introduces physics-informed masking\tkimuradel{ strategies to improve} for MAE pretraining.
Beyond learning techniques, alternative sensing embedding encoder architectures of FMs have been explored, such as frequency transformers~\cite{kara2024freqmae}, mixer models~\cite{ekambaram2023tsmixer, chentsmixer}, and recently emergent state-space models~\cite{shams2024ssamba, bhethanabhotla2024mamba4cast, zeng2024c}.

\subsubsection{Multimodal Models.}~FMs\tkimuradel{for multimodal sensor embeddings} that integrate heterogeneous signals from a fixed set of sensor modalities into joint embeddings have emerged for multimodal CPS-IoT applications~\cite{narayanswamy2024scaling,thapa2024sleepfm,mo2024iot, VibroFM}. 
\texttt{Cosmo}~\cite{ouyang2022cosmo} enhances modality representations with augmented fusion for contrastive learning.
\texttt{Cocoa}~\cite{deldari2022cocoa} improves cross-modal coherence with discriminative objectives on temporally distant samples.
\texttt{FOCAL}~\cite{liu2023focal} factorizes the representation into orthogonal shared and private subspaces to capture complementary and shared modality information.
Parallel works have explored aligning modalities to handle\tkimuradel{samples with missing modalities} missing data.
\texttt{Imagebind}~\cite{girdhar2023imagebind} binds images, text, audio, depth, thermal, and IMU data into a joint embedding space using only image-paired data.
\texttt{Babel}~\cite{dai2024advancing} proposes expandable\tkimuradel{ modality} networks to incrementally integrate\tkimuradel{ sensor} modalities.
{Similarly, }\texttt{MMBind}~\cite{ouyang2024mmbind} constructs pseudo-paired data from incomplete multimodal sources for\tkimuradel{ effective} multimodal pretraining.

\subsubsection{Flexible Modality Models.}~Existing works have explored FMs that adapt to flexible modalities, as available sensor modalities in CPS-IoT applications often vary\tkimuradel{ over time or across deployments} due to resource constraints, platform heterogeneity, and operating environment.
One approach involves models that infer modality availability at test time.
For example, missing modality can be indicated with special input value~\cite{srinivas2020training} or with a mask vector~\cite{qian2023contrastive}.
Transformers can use prompt tokens (similar to position embeddings) to mark sensor modality and adapt to missing data~\cite{lee2023multimodal,liang2022high,tsai2019multimodal}.
Models can also handle missing modalities at test time by performing probabilistic modality dropout or cross-modal attention masking during training~\cite{ma2022multimodal}.
Another approach trains reconstruction models with modality dropout to generate missing data from the available modalities~\cite{woo2023towards}. 
Alternatively, models like \texttt{Perceiver}~\cite{jaegle2021perceiver} adopt new architecture to handle arbitrary configurations of modalities.
Lastly, FMs for univariate time-series forecasting~\cite{ansari2024chronos, jin2023time, liu2024unitime} treat all sensor data as timestamped, scaled real-numbered values and can adapt to arbitrary modalities by handling input domain shift with scaling training data, though often unsuccessfully~\cite{mulayim2024time}.

\subsection{CPS-IoT FMs for Perception: A Task Perspective}
\noindent
The trend towards increasing homogenization of model functionality in broader AI as reflected in LLMs is also influencing CPS-IoT FMs resulting in an evolution from unifying sensor modality representations to unifying CPS-IoT tasks.
Below we analyze the SOTA in CPS-IoT FM from the perspective of breadth of tasks they handle, organizing them into four groups: fixed task, design-time configurable task, run-time selectable task, and run-time specifiable task.



\subsubsection{Fixed Task Models.}~These models perform a single specific task but can do so on a broad spectrum of sensor data types either in a zero-shot manner or with some fine-tuning for performance. The most prominent example of such FMs is time-series forecasting models, often referred to as Time Series FMs (TSFMs) despite their single-task focus, that have recently emerged in academic literature and commercial offerings. Examples include Amazon's \texttt{Chronos}~\cite{ansari2024chronos}, a family of open-source probabilistic forecasting models based on the T5 architecture~\cite{raffel2020exploring}, \texttt{Moirai}~\cite{woo_unified_2024}, which employs an encoder-only transformer for multivariate time series, \texttt{LagLlama}~\cite{rasul_lag-llama_2024}, which leverages the decoder-only Llama architecture~\cite{touvron2023llama} for probabilistic forecasting, and \texttt{TimesFM}~\cite{das2023decoder}, which uses a decoder-based transformer with residual blocks for multivariate time-series forecasting.


\subsubsection{Configurable Task Models.}~FMs focused on representing raw sensor data into embedding vectors fall under this category, such as those discussed in the preceding subsection (e.g., \texttt{LIMU-Bert}~\cite{xu2021limu}, \texttt{FOCAL}~\cite{liu2023focal}, \texttt{TimesNet}~\cite{wu2022timesnet} etc.). Pretrained via SSL on a broad spectrum of data, they can be configured for specific tasks by pairing with a downstream stage. Task-specific labeled sensor data is used to train the downstream stage, which may also fine-tune the embedding FM. For many tasks, the downstream stage can be as simple as a linear probe, though more sophisticated models are often used. An excellent example is \texttt{MOMENT}~\cite{goswami2024moment}, a family of open-source FMs for time-series analysis. It serves as a versatile representation learning FM that, when paired with a fine-tuned linear probe, can handle tasks like imputation, anomaly detection, and long-horizon forecasting, or, with a fine-tuned classification head, can perform classification tasks.



\subsubsection{Selectable Task Models.}~
Selectable CPS-IoT FMs provide greater flexibility by allowing users to choose from a predefined set of tasks at run-time without requiring \textit{any} additional fine-tuning. These models use a metadata channel to specify the desired task from the fixed set. The \texttt{MOMENT} FM mentioned above can also function as a selectable task model, performing anomaly detection, imputation, short-horizon forecasting, and classification in a zero-shot manner without parameter updates.
Similarly, \texttt{TimeGPT}~\cite{garza2023timegpt} utilizes an encoder-decoder structure to perform zero-shot forecasting and anomaly detection without additional training. \texttt{UniTS}~\cite{gao2024units}, a unified multi-task time-series model, processes input data as tokens and employs a shared architecture for tasks like forecasting, classification, anomaly detection, and imputation. At deployment, prompt tokens—learnable embeddings fine-tuned for specific datasets or tasks—are appended to input tokens to provide task-specific context, enabling adaptation to new tasks or datasets without modifying the frozen pretrained model.

\subsubsection{Run-time Specifiable.}~Run-time specifiable tasks enable users to define new tasks at run-time via input specification (e.g., text). 
In NLP, text-to-text models\tkimuradel{incorporate task descriptions into the input text, using approaches like} use task-specific prefixes~\cite{raffel2020exploring} or text description in modern LLMs~\cite{brown2020language} for task definitions.
In the CPS-IoT domain, recent work leverages LLMs’ internal pretrained world knowledge to ingest task descriptions and exhibit emergent reasoning, planning~\cite{gundawar2024robust}, and optimization~\cite{yang2024large} abilities.
The key difference\tkimuradel{ across works} lies in whether input data (e.g., sensor measurements) and outputs (e.g., state estimates, control actions) are textually embedded\tkimuradel{ within input prompts and output}, or mapped via\tkimuradel{ specialized} adapters to or from the embedding space.
Examples of the first approach include \texttt{PromptCast}~\cite{xue2023promptcast}, \texttt{LLMTime}~\cite{gruver2024large}, \texttt{IoT-LLM}~\cite{an2024iot}, \texttt{Penetrative AI}~\cite{xu2024penetrative} and \texttt{LLMSense}~\cite{ouyang2024llmsense}.
These works demonstrate that LLMs, with their world knowledge and reasoning abilities~\cite{mirchandani2023large, lu2022frozen}, excel in zero- and few-shot settings, handling tasks from forecasting to complex inference that go beyond mere sequence processing and pattern matching~\cite{gu2024your}.
The second approach, such as \texttt{IoT-LM}~\cite{mo2024iot} and \texttt{Time-LLM}~\cite{jin2023time}, trains adapters to align a CPS-IoT system's input, output, or intermediate variables to the backbone LLM's semantic space, outperforming the first approach but requiring more complex design and fine-tuning.
Lastly, some recent works~\cite{zhang2023unleashing, zhang2024unimts, yan2024language} have explored leveraging label semantics for sensing classification without any task-specific training.

\subsection{Beyond Perception}
\label{subsec:beyond_perception}
\noindent
While existing research towards CPS-IoT FMs has primarily focused on encoding sensory data for perception, there is incipient work on FMs for other stages of the PCCA loops with the emergence of task- and platform-specific AI/ML models.
The successes of deep reinforcement learning in robotics~\cite{haarnoja2018learning}, autonomous vehicles~\cite{kiran2021deep}, drones~\cite{wang2019autonomous}, pan-tilt-zoom cameras~\cite{sandha2023eagle}, and building energy and HVAC management~\cite{yu2021review} catalyzed interest in FMs adaptable to deployments (e.g., different buildings, robotic bodies, etc.) without requiring additional training.
Particularly, the robotics community has a burgeoning body of FM research~\cite{firoozi2023foundation, hu2023toward} developing general-purpose robots and models generalizable across new platforms.
These FMs are often trained as vision-language action (VLA) models~\cite{ma2024survey, brohan2023rt, kim2024openvla}, or finetuned Vision Language Models (VLMs), to learn policies for tasks like manipulation and navigation from massive vision-language demonstration data with diverse robots~\cite{o2023open} and can be used for variety of tasks such as planning~\cite{song2023llm} in the PCCA loops of robots.
Alternatively, LLMs and VLMs are increasingly used via prompting as FMs for CPS-IoT tasks in robotics~\cite{vemprala2023chatgpt, driess2023palm}, HVAC control~\cite{sawada2024office}, and industrial control~\cite{song2023pre}, as well as tools for generating control code~\cite{vemprala2023chatgpt, liang2023code}, annotating sensor data~\cite{hota2024evaluating}, tasking CPS-IoT resources~\cite{liu2024tasking}, managing sensor privacy~\cite{wang2024privacyoracle}, and Q\&A over sensor data~\cite{imran2024llasa}.

%% file: sections/3-Experience.tex
\section{New Insights from Our Research} 
\label{sec:our-insights}
\noindent
Given the quickly expanding landscape of FMs for CPS-IoT applications, an interesting question becomes: Are FMs for the CPS-IoT domain simply an adaptation of more general research in AI/ML to new datasets and tasks (such as the approaches summarized in the previous section) or are there fundamental differences in how we should think of FMs for CPS-IoT applications? Several distinguishing characteristics of the CPS-IoT domain impact FM design: 

{\em 1. Tight resource/quality trade-offs:\/} CPS-IoT applications significantly shift acceptable resource/quality trade-offs for practical deployment of AI/ML. For example, recent statistics indicate that the cost of converting a standard car to an autonomous one is significantly higher than the cost of the original vehicle~\cite{day2019accelerating}. This is unacceptable and would hinder the pervasive use of mobile intelligence.

{\em 2. Spatial embodiment:\/} CPS-IoT applications are concerned with world state that evolves in physical time and space. Sensors sample that state from various vantage points with a range of observation modalities. FMs must abstract away from sensor properties, modalities, and locations to representations of the observed phenomena in space and time.

{\em 3. Historical context:\/} Historical context is critical to interpreting current phenomena. Today's LLMs are limited in the amount of context they can ingest. While this amount may be sufficient for understanding large passages of text, sensors can be sampled at much larger rates, with arbitrarily old events playing a significant role in interpreting the present. Novel solutions are needed to preserve important context.

{\em 4. Structural constraints:\/} The physical embodiment of CPS-IoT applications into a world that abides by laws of nature entails the existence of significant amounts of prior knowledge, including laws of physics, system dynamics, and other constraints imposed on the data at multiple levels of abstraction. FMs, in a way, are an expression of the underlying knowledge in a domain. Thus, training and/or inference should allow explicit representation and exploitation of domain constraints to improve reasoning efficiency.

Motivated by the above differences, we discuss a preliminary exploration of challenges arising from these distinguishing properties, solution requirements to address these challenges, and remaining open issues. 

\label{sec:our_research}
\input{sections/3-Experience-ResourceChallenge}

\input{sections/3-Experience-Spatiotemporal}
\input{sections/3-Experience-ComplexEvent}

\input{sections/3-Experience-Knowledge}

%% file: sections/3-Experience-ResourceChallenge.tex
\subsection{Tight Resource/quality Trade-offs} 
\label{sec:resource_quality_tradeoffs}


\begin{figure}[tbp]
    \centering
 \setlength{\abovecaptionskip}{0.cm}
    \setlength{\belowcaptionskip}{0.cm}
    \begin{subfigure}[b]{0.9\columnwidth}
        \includegraphics[width=\textwidth]{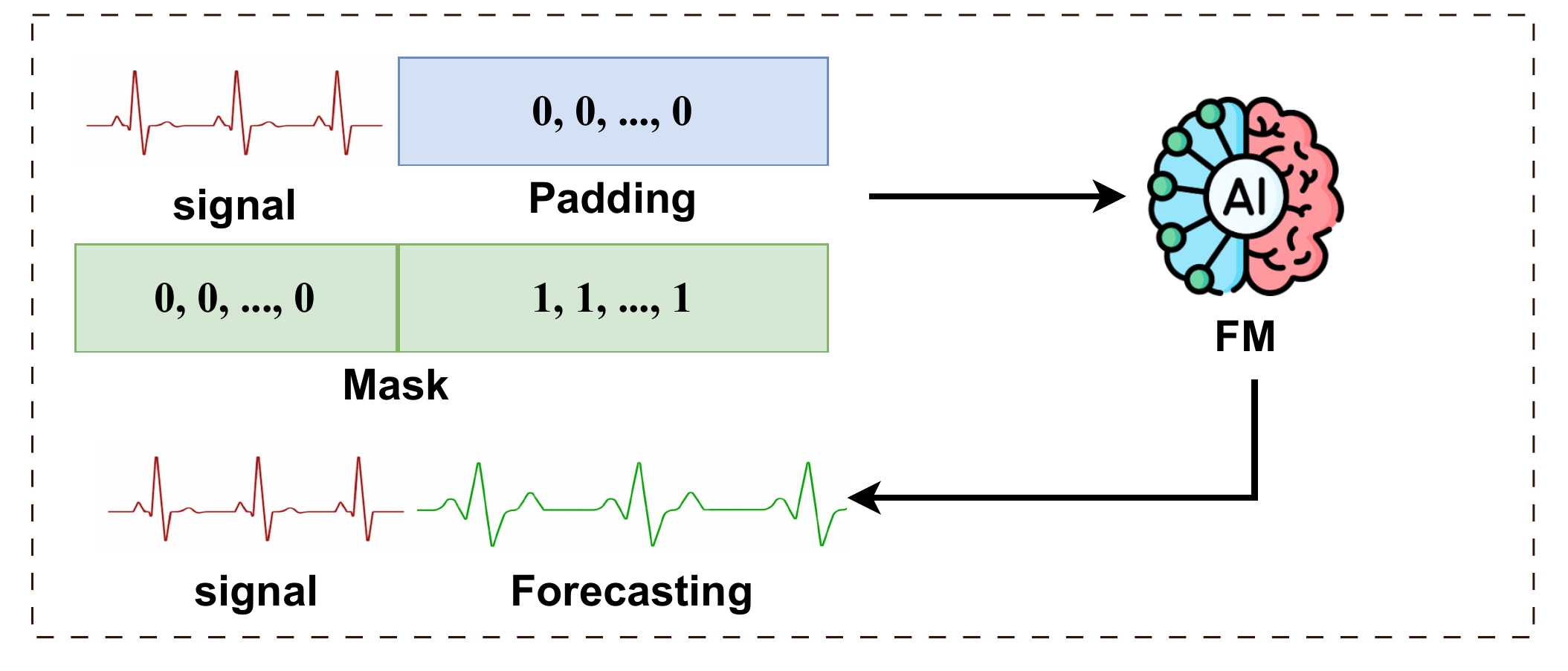}
    \end{subfigure}\\
    \begin{subfigure}[b]{0.9\columnwidth}
        \includegraphics[width=\textwidth]{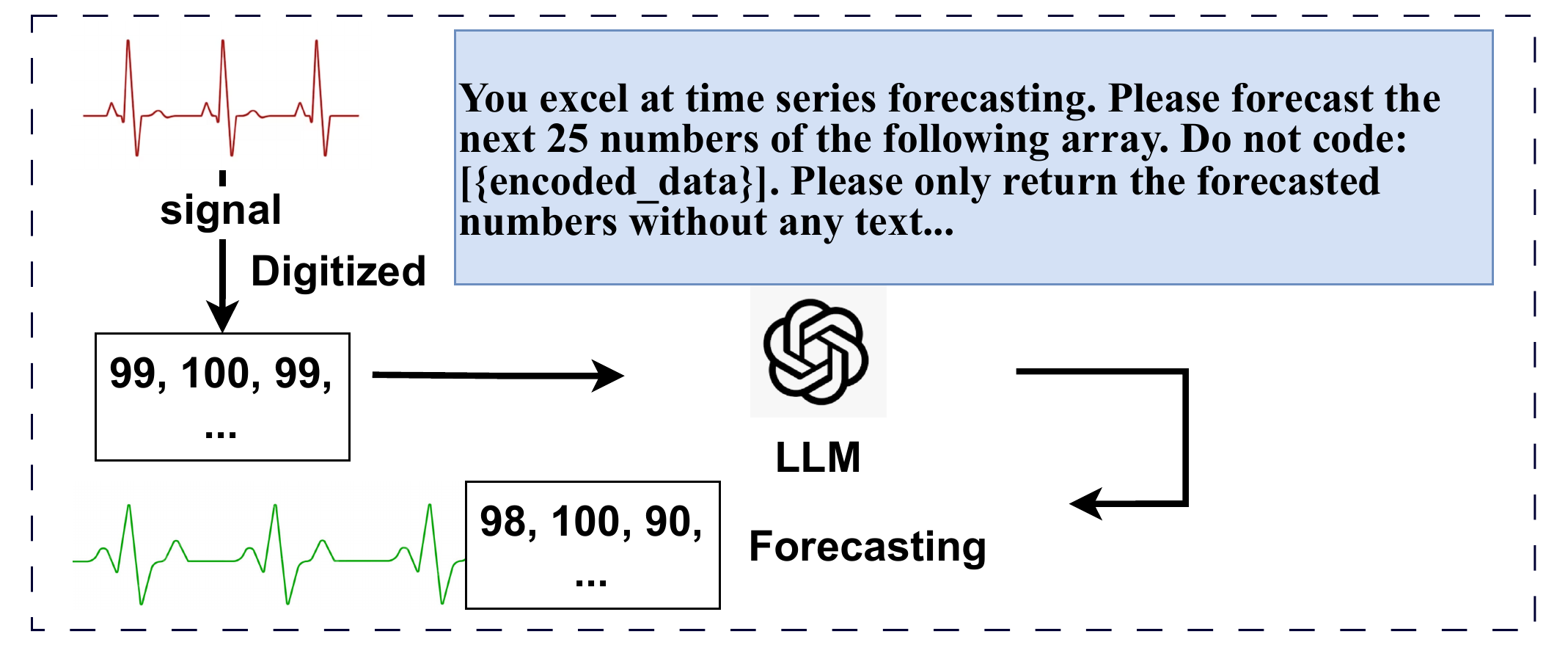}
    \end{subfigure}
    \caption{Test settings. Top: TSFM. Bottom: LLM.} 
    \label{fig:setting_FMsLLMD}
    \vspace{-4mm}
\end{figure}

\noindent
The integration of future CPS-IoT FMs with private sensor data and user context will favor implementations where the intelligence resides on the edge and not in the cloud. To what extent are improvements required in resource efficiency and inference quality for FMs to run on the mobile device?

\subsubsection{A Preliminary Experiment.}~
To address the above question, we test two existing general FMs on task-specific IoT data: a time-series FM (TSFM), MOMENT (of which we use two different sizes), and an LLM, Llama-3.2, with 1 billion parameters. Figure \ref{fig:setting_FMsLLMD} illustrates the experimental setup for these models. To assess their effectiveness, we consider two common mobile IoT tasks: extrapolation and imputation, executed on electrocardiogram (ECG) and photoplethysmogram (PPG) signals, downsampled to 50Hz, inspired by mobile health monitoring applications. We also compare these approaches to two classical baselines (variations of ARIMA models using \cite{pmdarima}). The first baseline, called ARIMA-pre-training (ARIMA-PT), fits an ARIMA model using an 80-20 training-evaluation split, after which the model remains fixed for evaluation. The second approach, termed ARIMA-online-training (ARIMA-OT), leverages the lightweight nature of ARIMA models to perform test-time fitting. 
We implement the models on a Google Pixel 8 Pro based on an open-source Java library \cite{arimajava} and the ExecuTorch framework \cite{executorch}.
Memory footprint is assessed by measuring the memory usage change before and after loading and running the model. Cold-start latency is defined as the total duration encompassing both the model loading phase and the execution of a single inference, while warm-start latency measures the time taken for a single inference after the model is already loaded. 

\begin{table}[b]
\centering
\caption{Accuracy, latency, and memory footprint of models. In the latency column, we evaluate both cold-start latency (left) and warm-start latency (right).}
\resizebox{\columnwidth}{!}{%
\begin{tabular}{ccccccc}
    \hline
    \textbf{Model} & \bm{$ECG_{Imp}$} & \bm{$ECG_{Ext}$} & \bm{$PPG_{Imp}$} & \bm{$PPG_{Ext}$} & \textbf{Latency (ms)} & \textbf{Mem. (MB)} \\ \hline
    
    ARIMA-OT & 0.1387 & 0.1276 & 0.0083 & 0.0070 & 42.8/11.3 & \textbf{0.33} \\
    ARIMA-PT & \underline{0.1195} & 0.1235 & 0.0077 & 0.0078 & \underline{2.0}/\underline{1.6} & \underline{0.74} \\
    GridARIMA-OT & 0.1263 & \underline{0.1157} & 0.0102 & 0.0071 & 148.7/40.6 & 2.30 \\
    GridARIMA-PT & 0.1562 & 0.1269 & \underline{0.0072} & \underline{0.0068} & \textbf{1.8}/\textbf{1.3} & 12.49 \\
    Moment-S & 0.1442 & 0.1392 & 0.0081 & 0.0075 & {217.0}/72.5 & {88.7} \\
    Moment-L & \textbf{0.1030} & \textbf{0.1134} & \textbf{0.0050} & \textbf{0.0061} & 2359.4/618.2 & 1207.7 \\
    Llama3.2-1B & 1.9052 & 3.3038 & 4.0773 & 6.1396 & 9967.5/7641.7 & 1662.7 \\
    \hline
\end{tabular}
}
\label{table:exp_FMsLLM}
\end{table}
\subsubsection{Experimental Results and Recommendations.}~Table~\ref{table:exp_FMsLLM} shows that the largest of the time-series FMs, MOMENT-L did outperform both versions of ARIMA on both tasks. However, this was accomplished at the expense of 2-3 order of magnitude increase in latency, and over 3 orders of magnitude increase in memory footprint. 
For example, Moment-L required 1.2 GB of memory and incurred a cold-start latency of 2.3 seconds, compared to 0.74 MB and 2ms cold-start latency for ARIMA-PT. Moreover Llama failed to outperform ARIMA, despite its higher resource consumption than even MOMENT-L. This observation highlights the limitations of today's general LLMs in processing low-level sensory data directly, suggesting that their strengths lie in handling metadata, high-level reasoning, or language-based analytics, rather than raw time series analysis. 

The experiment motivates developing a new generation of FMs specifically designed and optimized for CPS-IoT applications of interest. Such FMs, if successful, might become new architectural abstractions -- a part of the mobile OS -- that allow handling myriads of user-specific intelligent tasks. But first, one must understand more carefully how CPS-IoT applications differ and what these differences imply in terms of FM design and customization.

%% file: sections/3-Experience-Spatiotemporal.tex
\subsection{Spatial Embodiment}
\label{sec:spatial}
\noindent
CPS-IoT applications distinguish {\em world state\/} (of monitored phenomena) from the {\em observed\/} (or {\em measured\/}) views as sampled by sensors at specific vantage points. There is usually the notion of a {\em channel\/} that propagates state from the source phenomenon to the observers, incurring various distortions along the way. This creates an inconsistency: Ingested FM data are {\em sensor data\/} representing distributed local observations. The goal, however, is to represent the actual {\em world state\/}. For example, in acoustic monitoring, one may want to recover the real sound, location, and speed of a moving object from projections perceived by individual sensors at different vantage points. CPS-IoT FMs must therefore abstract from the time-series data of individual sensors to representations of the underlying observed phenomena. 
How can the training of FMs be nudged to represent the observed environment, not sensor data? Said differently, how to ensure the invariance of the representation from the observation instruments and their locations? 
\subsubsection{Challenges.}
~To allow FM training to decouple {\em local signal projections\/} from the unified latent semantic representation of the observed phenomena, it needs to understand the {\em spatial structure\/} of observations. Arbitrarily placed sensors simply perform an irregular local sampling of the scene. Their positions must be encoded by the FM to understand how the scene is sampled.  
Traditional positional embeddings, such as those common for text and images, are typically designed to encode the relative order of input tokens in sequential or grid-based data.
In contrast, sensors are not deployed in regular lines or grids. 
Thus, in multi-view or multi-vantage sensing applications, sensor geo-location embedding (that directly encodes the specific location information of each vantage point) is desirable.
Location embeddings should further generalize to differences in absolute positioning that do not impact spatial reasoning.
For example, they can encode relative distances, not absolute GPS locations. 
Alternatively, they can use augmentations that abstract away the absolute locations to ensure that the model focuses on relative positioning features.
In addition to the geo-location embedding,  timestamp embeddings should also be incorporated to specify the temporal metadata of observations. 
A multi-vantage model must also understand the relations between vantage points. For example, in vision, geometric transformations and/or neural network-based approaches (e.g., NeRF~\cite{mildenhall2021nerf}) can be applied to construct an image from a new angle or vantage point given existing images from other vantage points.
To attain the same effect with arbitrary sensors, one training approach could be to mask a portion of a sensor's time series and reconstruct it from signals at other vantage points, forcing the model to learn signal dependencies on location.

\subsubsection{A Preliminary Experiment.}
\begin{figure}[htb]
    \centering
    \includegraphics[width=1.0\linewidth]{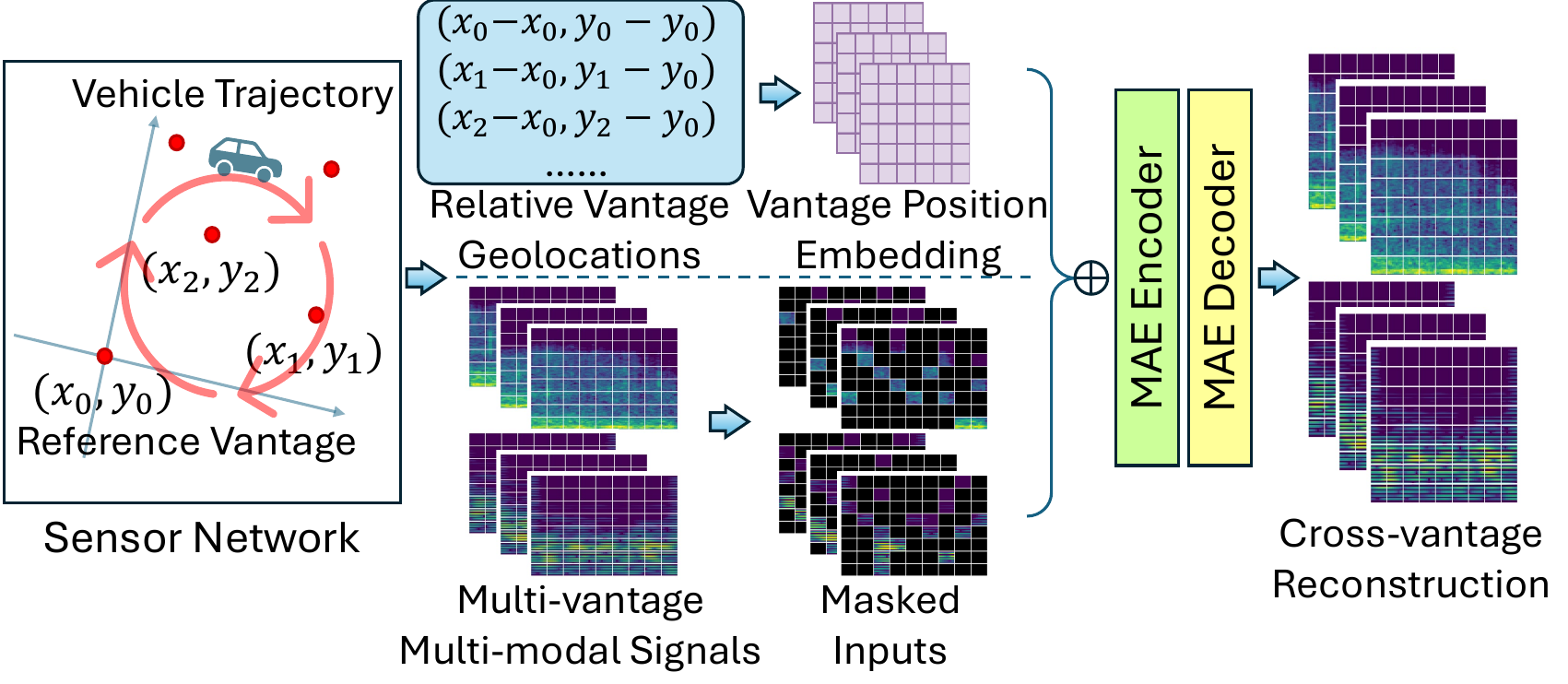}
    \vspace{-0.5cm}
    \caption{An architecture for spatial representation learning.}
    \label{fig:architecture}
    \vspace{-4mm}
\end{figure}

~For a proof of concept, we used a multimodal MAE-based framework that consists of an encoder and a decoder. Input signals from different vantage points are transformed into visual representations (namely, spectrograms). A dedicated encoder processes each modality, and the resulting embeddings are combined in a joint encoder to enable a cross-modal representation. 
The overall architecture is illustrated in Figure ~\ref{fig:architecture}. 

To encode positional information, the model employs a 3-dimensional learnable positional embedding that integrates time, frequency, and spatial vantage point information. These embeddings are added to the masked input before being processed by the encoder. Notably, relative coordinates of vantage points are encoded instead of absolute locations to ensure the translation invariance of the spatial positional embeddings. To uncover spatiotemporal dependencies, the raw inputs are divided into patches, with a subset selectively masked across specific vantage points, time steps, or frequency bands. The auto-encoder is then trained to reconstruct the complete stacked spectrograms. This reconstruction process enables the framework to extract spatiotemporal relations in a self-supervised manner by inferring signals across related vantage points, times, and frequency bands.

\subsubsection{Experimental Results and Recommendations.}
\begin{figure}[t!]
    \centering
    \includegraphics[width=\columnwidth]{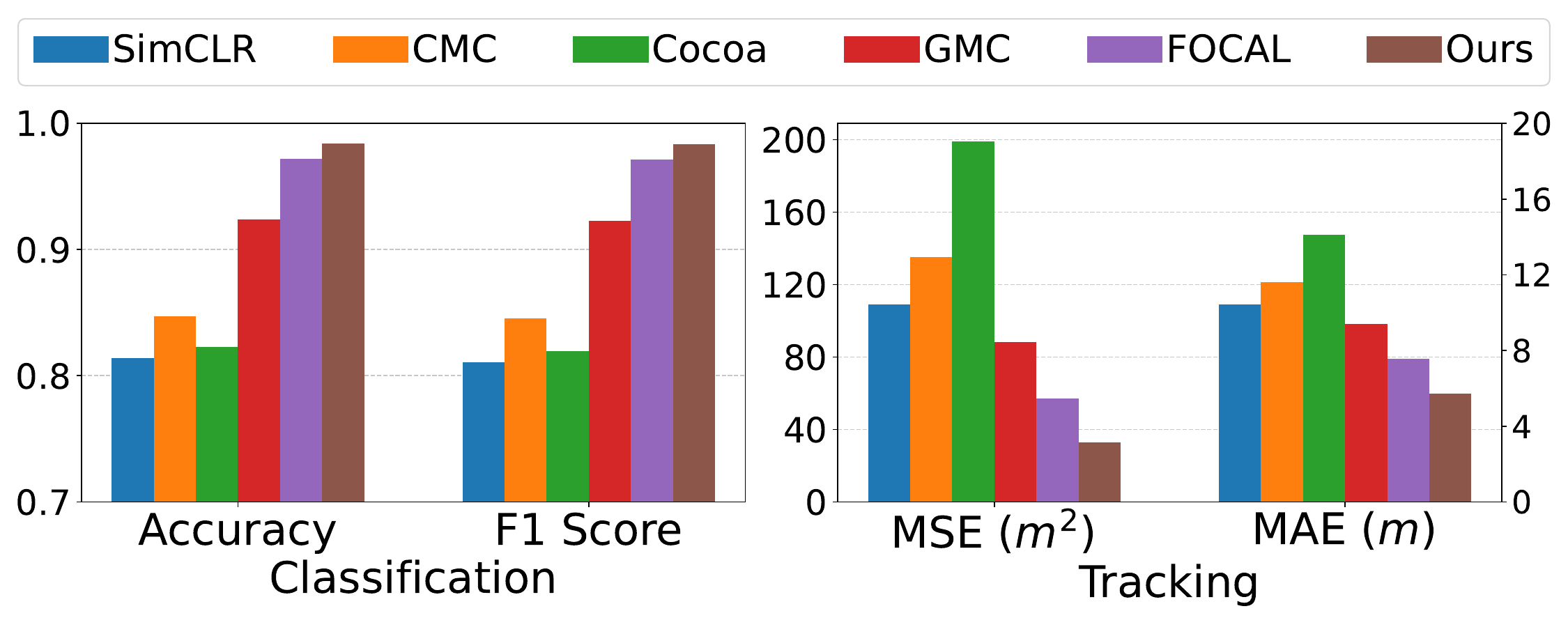}
    \vspace{-8mm}
    \caption{Performance comparison of multi-node classification (measured by accuracy $\uparrow$ and F1 score $\uparrow$) and tracking (measured by MSE $\downarrow$ and MAE $\downarrow$) tasks.}
    \label{fig:metrics}
    \vspace{-6mm}
\end{figure}

~ 
%
%
%
To test if the newly-trained model can recover physical states of observed phenomena, we use it to perform {\em target tracking\/}. Unlike classification, where data from any one sensor might be sufficient for identifying the target, {\em tracking\/} requires collective use of multiple sensor signals and positions to learn the equivalent of {\em localization\/}. 
We evaluate it on a self-collected dataset using a network of six Raspberry Pi nodes strategically positioned along vehicle routes to collect acoustic and seismic signals.
During the data collection, civilian vehicles were driven by the sensors at varying speeds and directions, with the vehicle's real-time GPS coordinates recorded for ground-truthing purposes.
%
%
We present the trained model's classification and tracking performance in Figure~\ref{fig:metrics} compared to state-of-the-art contrastive baselines~\cite{chen2020simclr, tian2020contrastive, deldari2022cocoa, poklukar2022gmc, liu2023focal}. 
Our method consistently achieves the highest accuracy and F1 score for classification and the lowest Mean Squared Error (MSE) and Mean Absolute Error (MAE) for tracking. 

The experiment demonstrates the importance of investigating novel approaches to FM pretraining that nudge the FM to learn representations of {\em physical world state\/}, as opposed to merely encoding the projections of that state, comprising sensory time-series signals. This is often called the {\em inverse problem\/}~\cite{tarantola2005inverse}, as distinguished from forecasting future observations. Many challenges must be solved to develop general mechanisms for distilling unified {\em sensor-agnostic latent representations of environmental state\/} (collected from a wide range of sensing modalities and vantage points), and mechanisms for automating {\em state interpretation\/}, given world knowledge. The problem is more challenging in cluttered environments, calling for solutions that augment sensor measurements with prior knowledge to disambiguate among multiple competing inverse hypotheses regarding complex state from limited available observations. The final representation should also be generalized across different sensor configurations, abstracting away from specifics such as sensor calibration details, gains, and sampling rates. 

%% file: sections/3-Experience-ComplexEvent.tex
\subsection{Historical Context} 
\label{sec:historical}
\noindent
The importance of temporal events (that extend over time) to the interpretation of observed physical phenomena makes historical context of present observations an important part of their semantic representation.  
Most existing work on CPS-IoT applications has focused on short-time perception tasks, such as human activity recognition or object detection, which typically require only a few seconds of sensor data for inference. However, to achieve a human-like understanding of the world, a system must capture high-level contextual information over extended periods of time, an aspect often overlooked in current work. 

\begin{figure*}[t]
    \centering
\includegraphics[width=0.9\textwidth]{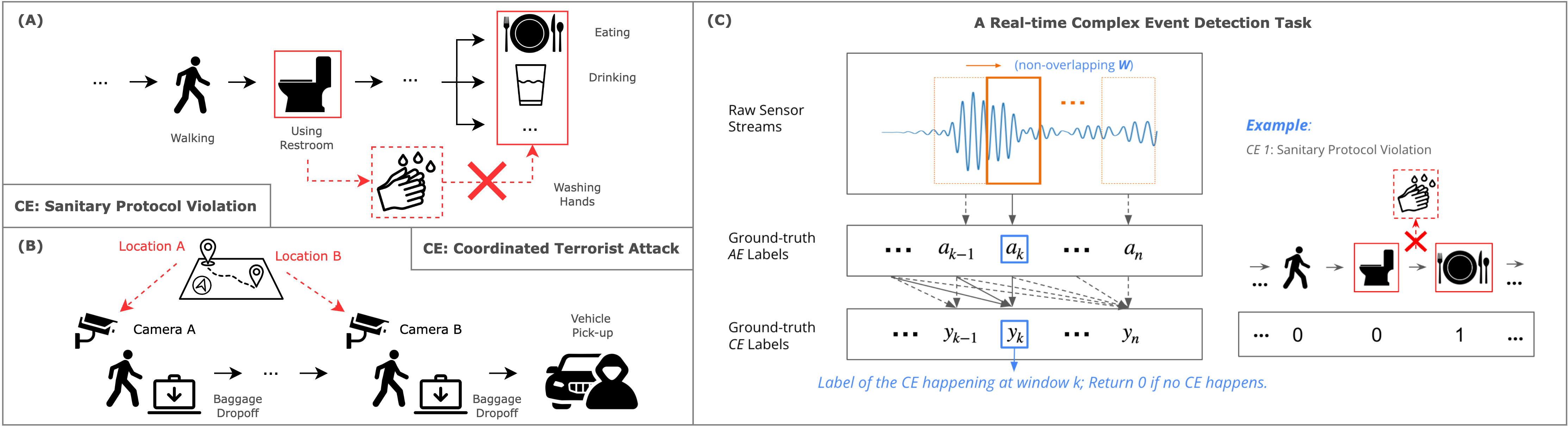}
    \caption{(a) Sanitary protocol violation in smart home health monitoring system. (b) Detecting coordinated terrorist attacks at different locations across the city using the surveillance system. (c) In a real-time complex event detection (CED) task, only the raw sensor streams and ground-truth complex event labels are provided.}
    \label{fig:ce_overview}
    \vspace{-3mm}
\end{figure*}

For example, for an outside observer to ascertain that an office building is empty (say, at the end of a work day), it needs the accumulated record of all entry and exit events since the building was last unlocked. Considering a camera with a 30 frame/second rate, an 8-hour work-day generates nearly one million frames, far more than the context window of modern LLMs. It is therefore important to know {\em what needs to be remembered\/}. Unfortunately, the key events to remember depend on the situation. To describe different situations, it is useful to define the concept of a \textit{complex event}. A complex event represents a high-level scenario with spatiotemporal rules and patterns that require aggregating and reasoning over numerous short-term activities, which we call \textit{atomic events}. 
For example, Fig.~\ref{fig:ce_overview}(a) defines a complex event where an intelligent assistant on a mobile device understands a sanitary protocol and alerts users of potential violations. Fig.~\ref{fig:ce_overview}(b) shows a scenario in a smart facility, where a surveillance system\tkimuradel{must } detects potentially suspicious activity, such as an unusual parcel hand-off, by analyzing data across distributed cameras. 

\subsubsection{Challenges.}
~Complex event detection (CED) introduces several challenges. First, 
the pattern of interest (e.g., the trained self-attention matrix of the FM's encoder) must identify key atomic event occurrences relevant to the complex event while ignoring irrelevant activities. Let us call the latter, "don’t care" elements, or "X." For example, the sanitary protocol can be represented as "Use restroom → X → Wash hands → X → Eat," where "X" includes other irrelevant activities like "walking" or "sitting." Incorporating "X" broadens the range of possible matching sequences, and the temporal duration further amplifies this space exponentially. Second, complex events have much longer time dependencies. In the sanitary protocol example, violation occurs when the person skips "Wash hands" after "Use restroom" and before "Eat". However, the time gap between those atomic events can be significant.
Third, complex events often require immediate attention. For instance, in a nursing monitor system, we shouldn't wait to analyze data until the end of the day. Instead, we need immediate alerts when safety is violated.

\begin{table}[htb]
\caption{Evaluation results of LLMs.}
\vspace{-3mm}
    \begin{center}
    \small
    \setlength\tabcolsep{1.5pt}%
    \begin{tabular}{@{}lccccccccccc@{}}\toprule 
    & \multirowcell{2}{\makecell[c]{$Length$\\$Acc.$}}  && \multicolumn{4}{c}{$Coarse$ $F1$}  && \multicolumn{4}{c}{$Conditional$ $F1$}\\
    \cmidrule{4-7} \cmidrule{9-12}
    \textit{Zero-shot} & && $e_1$ & $e_2$ & $e_3$ & Avg.&&  $e_1$ & $e_2$ & $e_3$ & Avg.\\ 
    Qwen2.5-7B & 0.04 && 0.51 & 0.67 & 0.67 & 0.62 && 0.0 & 0.18 & 0.05 & 0.07 \\
    Qwen2.5-14B & 0.12 && 0.60 & 0.66 & 0.57 & 0.61 && 0.14 & 0.15 & 0.04 & 0.11\\
    GPT-4o-mini  & 0.04 && 0.0 & 0.0 & 0.64 & 0.21 && 0.0 & 0.0 & 0.0& 0.0 \\
    GPT-4o & 0.12 && 0.87 & 0.78 & 0.80 & 0.82  &&  0.14  & 0.59  & 0.03  & 0.25 \\
    \midrule
    \textit{Few-shot} ($k = 3$) & $Acc.$ && $e_1$ & $e_2$ & $e_3$ & Avg.&&  $e_1$ & $e_2$ & $e_3$ & Avg.\\ 
    Qwen2.5-7B & 0.04 && 0.49 & 0.68 & 0.71 & 0.63 && 0.0 & 0.19 & 0.04 & 0.08 \\
    Qwen2.5-14B & 0.14 && 0.62 & 0.66 & 0.60 & 0.63 && 0.13 & 0.13 & 0.03 & 0.10\\
    GPT-4o-mini & 0.03 && 0.0 & 0.0 & 0.58 & 0.19 && 0.0 & 0.0 & 0.0& 0.0 \\
    GPT-4o & 0.16 && 0.87 & 0.81 & 0.81 & 0.83 &&  0.13 & 0.63  & 0.14  & 0.30   \\
    \bottomrule
    \end{tabular}
    \label{tab:results}
    \end{center}
\end{table}

\subsubsection{A Preliminary Experiment.}~
To investigate the performance of existing LLMs on a real-time CED task, we designed a multimodal complex event dataset in a smart health monitoring setting, 
%
and created a stochastic simulator that mimics daily human behaviors and synthesizes the corresponding sensor traces using existing IMU and Audio datasets for the underlying atomic activities, WISDM\cite{wisdm} and ESC50\cite{esc50}. Fig.~\ref{fig:ce_overview}(c) illustrates the task.
To help the LLM, we assumed a perfect labeling tool that detects the ground truth text label of the atomic activity and passes it to the LLM. We then explained in the text prompt what constitutes a complex event.  We use three metrics to evaluate the LLM's performance. The
\textit{Length Accuracy} metric determines if the the complex event labels given by LLMs have the same length as the input sequence. The 
\textit{Conditional $F1$ Score} evaluates element-wise (time-wise) $F1$ score for three complex event labels, conditioned on the case when the LLM outputs a complex event sequence with the correct length $T$. 
\textit{Coarse $F1$ Score} is a sample-wise coarse $F1$ score that evaluates complex event labeling at high-level. It does not require a precise match between the predicted and ground-truth complex event labels at every timestamp. It only requires the LLM to recognize a complex event type in the 5-minute sample correctly.
Table~\ref{tab:results} shows the results. It can be seen that the LLMs performed somewhat poorly at complex event detection for all types of complex events considered.

\begin{figure}[!t]
    \centering
    \includegraphics[width=0.95\columnwidth]{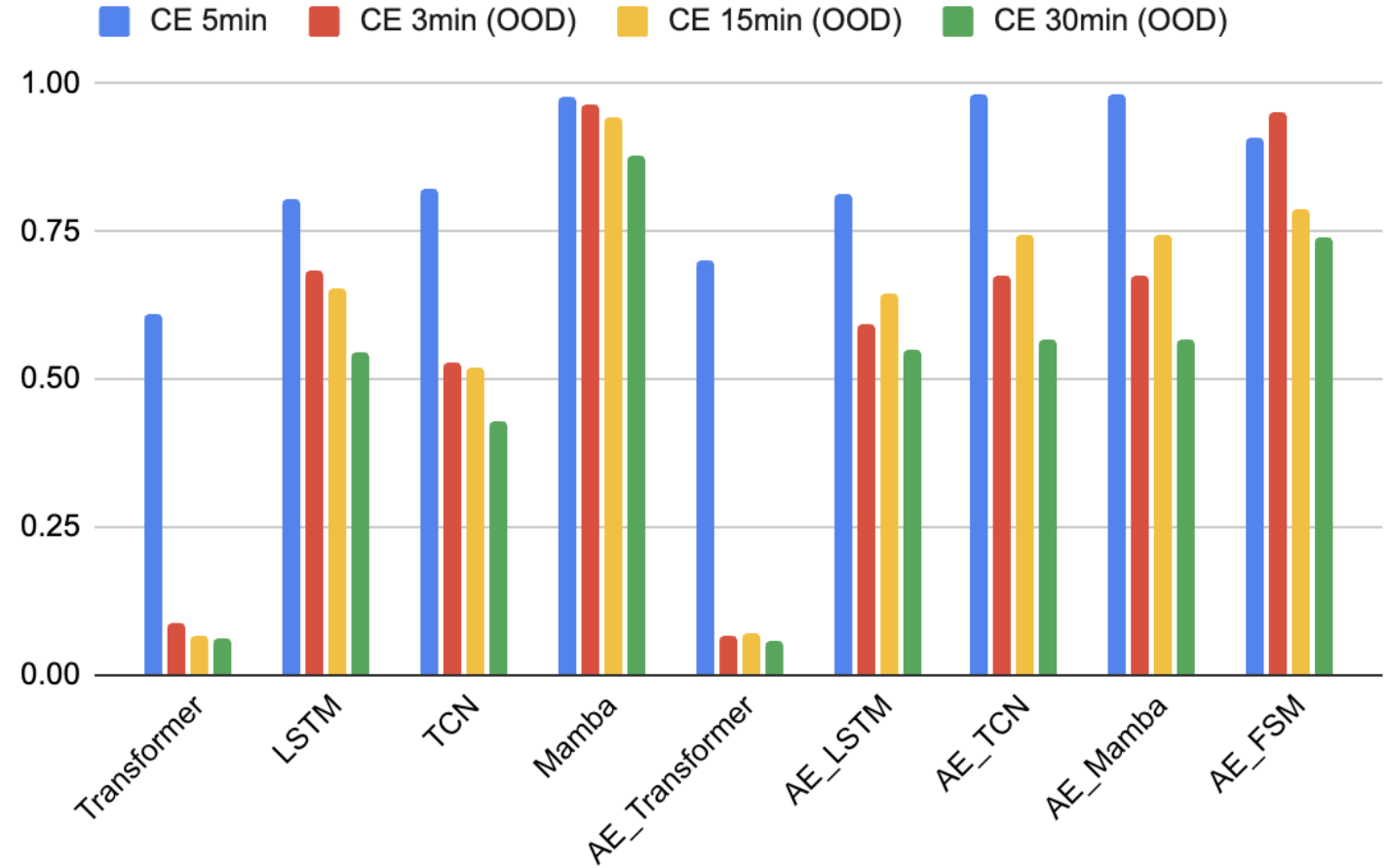}
    \vspace{-3mm}
    \caption{Average F1 scores of models on complex events with different temporal spans.}
    \label{fig:ce_train_results}
    \vspace{-7mm}
\end{figure}

Two approaches were then considered to building an improved real-time Complex Event Detection (CED) framework. The first uses sequential neural network models like CNNs, Transformers, and RNNs to learn complex event rules in a data-driven manner. The second employs {\em neurosymbolic\/} architectures with hand-coded rules, such as DeepProbLog\cite{deepproblog} and NeurASP \cite{neurasp}, utilizing neural network outputs for probabilistic symbolic computation. The neurosymbolic method we designed integrates neural networks for recognizing atomic activities with user-defined rules for complex event detection. 
%
We compare three model types for complex event detection: (1) \textit{End-to-End Models}: Includes Transformer, LSTM, TCN, and Mamba, trained on sensor latent embedding sequences and complex event labels; (2) \textit{Concept Bottlenecked Models}: Consists of AE + Transformer, AE + LSTM, AE + TCN, and AE + Mamba. These models use a pretrained classifier (AE) to obtain atomic activity labels, which are then used to train the neural backbones; (3) \textit{Neurosymbolic Model}: The AE+FSM model employs the most probable atomic activity label as input for rule-based finite state machines (FSMs).


\subsubsection{Experimental Results and Recommendations.} ~Figure~\ref{fig:ce_train_results} presents our preliminary results for various models. We also tested these models on out-of-distribution (OOD) complex events lasting 3 minutes, 15 minutes, and 30 minutes, all adhering to the same rules but varying in temporal span. The Mamba and AE+FSM models outperformed others on average. The AE + FSM model incorporates correct complex event rules; its performance declines greatly with longer traces due to cumulative errors from imperfect atomic event inference. While the Mamba model showed the best generalization on the OOD test sets, we still noted a performance drop as the temporal span increased. 

The above experiments show that FMs capable of reasoning over long and complex temporal patterns are still lacking. Though LLMs have potential to perform well on CED tasks, the current models still suffer from hallucinations and poor ability in long-chain reasoning. 
In our preliminary experiments on different model architectures and methods, we also find that state-based methods such as the Mamba model and FSM engines are more suitable when dealing with complex event patterns that may span a long time, as they can compress information into states efficiently. However, CED tasks remain challenging in many ways. For example, it remains to design efficient means to define new types of complex events, allowing the FM to detect them in a zero-shot of a few-short manner. Input tokenization is another important concern. With meaningful events occurring at multiple granularities, the best size for input time-series sensor data tokens is unclear. Another challenge is to scale context window size and develop mechanisms to efficiently remember old (but important) events in continuous and long sensor streams. In particular, the ability to incorporate human knowledge and the use of neurosymbolic architecture to guide attention seem like key architectural requirements.

%% file: sections/3-Experience-Knowledge.tex
\subsection{Structural Constraints} 
\label{sec:structural_constraints}
\begin{figure}[t!]
    \centering
    \includegraphics[width=\columnwidth]{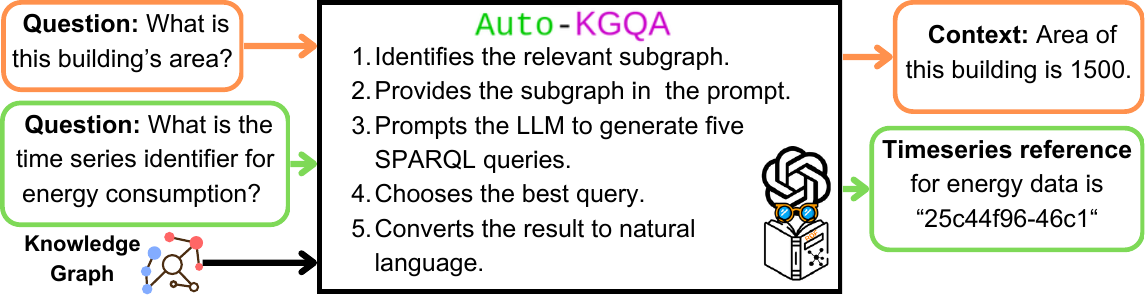}
    \caption{Example input outputs for our experimentation with AutoKGQA to extract contextual and sensory information.}
    \label{fig:kgqa}
    \vspace{-5mm}
\end{figure}
\noindent
To solve the inverse problem referred to in Section~\ref{sec:spatial}, define complex events referred to in Section~\ref{sec:historical}, or simply imbue an LLM with specialized domain-specific knowledge~\cite{zhang2024comprehensive}, one must express
structural constraints that encode any explicit prior knowledge about the world.
Knowledge graphs (KGs) are an ideal candidate to bridge human-like understanding of the world with the autonomous capabilities of FMs \cite{chen2023large}, particularly to enable complex and context-aware tasks in CPS IoT systems. The integration of LMMs and KGs has indeed been an active research topic~\cite{pan2024unifying}. Specifically, KGs can encode structured relationships and metadata that reflect the architecture and dynamics of the system, as well as describe the data-generating process. They essentially provide two key types of information: (1) contextual information, such as system characteristics, interdependencies, functional hierarchies, and dynamics; and (2) knowledge about the sensor data itself (i.e., metadata). 
The development of domain-specific KGs for CPS-IoT applications is thus an important endeavor. 

A particularly attractive solution is to couple KGs with LLMs so that arbitrary queries for the contextual and metadata information needed to address CPS-IoT tasks can be expressed in natural language, thus creating a sort of dynamic linker that can identify the relevant contextual knowledge needed to achieve arbitrary tasks at runtime. LLMs can be coupled with KGs in at least three ways: LLM-augmented KGs, which use existing pretrained LLMs to perform KG tasks such as Knowledge Graph Question Answering (KGQA, e.g.~\cite{avila_experiments_2024}); KG-enhanced LLMs which use KGs during pretraining and fine-tuning of the LLM; and, though less explored, synergized LLMs + KGs which use both methods to achieve mutually beneficial results \cite{li2024synergizing, pan2024unifying}. In general, there is a large body of work on LLMs leveraging KGs for a variety of applications~\cite{pan2024unifying} outside of CPS-IoT. 

\subsubsection{Challenges.}~
Selecting relevant sensory and contextual data for CPS-IoT applications is challenging due to the diversity and volume of time series data and complex system configurations. CPS-IoT systems can comprise of thousands or millions of sensors/actuators deployed over equally large number of components with exponentially many more relationships, leading to complex and massive KGs. The goal is to automatically retrieve and process relevant contextual information and time series data based on high-level task descriptions by leveraging semantic context and sensor relationships encoded in KGs.

\subsubsection{A preliminary Experiment.}~ 
To identify the methodological differences when applying such approaches to the CPS-IoT domain, we conduct a preliminary experiment using buildings as a use case. Figure \ref{fig:kgqa} illustrates our experimentation. Specifically, 
we examine a scalable LLM-based KGQA method for information retrieval in CPS IoT systems, focusing on sensory data~\cite{avila_experiments_2024}. The buildings domain is particularly suitable for this investigation due to the maturity of KGs to describe buildings and their CPS-IoT systems, such as the Brick schema \cite{balaji_brick_2016}. 
Thus, 
we used the KG of a real-world building (vm3a) from the Mortar repository \cite{fierro_mortar_2019}, which was created with the Brick Schema and contained 1,609 schema terms (T-box) and 61,366 instance-level resources (A-box). We explored the task of calculating the energy use intensity (EUI) of a building, which is determined by dividing a building's annual energy consumption (kWh) by its total floor area, requiring retrieval of (1) dynamic energy consumption time series and (2) the static building area value. Although, ideally, we would like FMs to take the text describing the full application/task directly, existing KGQA tools are designed for what is described as \textit{simple questions} \cite{huang2019knowledge}. Thus, to accommodate to this limitation, we divided the task description into two subqueries: (1) retrieving the total building area, and (2) identify the relevant energy consumption time series reference. This example is valuable because it involves extracting a time series reference (i.e., metadata about a dynamic data source) and building area (i.e., contextual information). Due to human ambiguity being a common challenge in KGQA , we paraphrased each input with varying levels of specificity, as shown in Table~\ref{table:query_questions}. For question answering, we employed AutoKGQA \cite{avila_experiments_2024}, a state-of-the-art framework recognized for its domain independence and capability to function without requiring any training/fine-tuning. We tested AutoKGQA with Llama 3.1 405B and GPT-4o, both yielding identical results shown in Table \ref{table:query_questions}. We set a similarity threshold of 0.15 and an LLM temperature of 0 to maximize consistency in the results.

\begin{table}[ht]
\centering
\caption{Questions with varying specificity for Building Area (1-5) and Timeseries ID queries (6-10)}
\vspace{-2mm}
\resizebox{\columnwidth}{!}{%
\begin{tabular}{l p{8cm} c} 
\hline
\bm{$Q_{id}$} & \textbf{Inputs} & \textbf{405B/4o} \\
\hline
1 & What is the value of the area of this building? & $\checkmark$ \\
2 & Retrieve the numerical value of this building's area. & $\times$\\
3 & What is the area of this building? & $\times$ \\ 
4 & Get the area details of a building. & $\times$ \\ 
5 & How large is this building? & $\times$ \\
\hline
6 & What is the timeseries ID of the timeseries reference of a building electrical meter? & $\checkmark$ \\
7 & Retrieve the unique timeseries ID associated with the timeseries reference of an electrical meter in this building. & $\checkmark$ \\
8 & What is the timeseries ID of a building electrical meter? & $\times$ \\
9 & Identify the ID for the data series connected to a specific electrical meter in the building. & $\times$ \\
10 & What is the timeseries for the energy consumption of this building? & $\times$ \\
\hline
\end{tabular}
}
\vspace{-4mm}
\label{table:query_questions}
\end{table}

\subsubsection{Experimental Results and Recommendations.}~ Beyond the established challenges for KGQA known in the literature, we have identified issues unique to CPS IoT systems, below.

\textit{One-to-Many Mapping:}  
Unlike traditional KGQA questions targeting clear objects (e.g., "Obama," "University of Edinburgh"), CPS concepts are often vague (e.g., "Energy"). LLMs struggle to map such inputs to relevant KG entities. For example, the FAISS relevance-based search for $Q_{id}=10$ returned 16 triples, including \texttt{Building}, \texttt{Energy\\\_Sensor}, and \texttt{Energy\_Zone}, but not \texttt{Building\_Electrical\\\_Meter}, which the KG used to associate the energy consumption time series. These design discrepancies, influenced by practitioner decisions, lead to variability in class availability, as noted in \cite{bennani_query_2021}. Traditional KGQA typically avoids this complexity due to the clear uniqueness of its subjects. A solution such as restricting reasoning to the A-box is insufficient when schema-level (T-box) reasoning, such as class hierarchy inference, is required. Balancing these needs remains a key challenge in adapting KGQA to CPS.

\textit{Size of the Knowledge Graph/Subgraph:}  
LLM-based KGQA approaches either input the entire KG or extract relevant subgraphs to fit token limits. However, we observed that extraction solely based on semantic similarity often exceeded token limits, as sensor-related inputs pull in all sensor-associated classes. This makes balancing efficiency and relevance difficult. When we tested GPT-4o with prompting relevant subgraphs, they successfully generated all SPARQL queries from the given questions, highlighting that the main challenge lies in supplying LLMs with the appropriate relevant subgraph.

\textit{Relation Linking:}  
Previous KGQA efforts have attempted to determine the number of hops required from the description of a question (e.g., identifying the zip code of the capital city of China involves two hops)\cite{wu_retrieve-rewrite-answer_2023}. Building ontologies requires more complex relation linking since users often lack awareness of system configurations. For example, finding the timeseries ID of a building electrical meter ($Q_{id}=8$) may seem like a single-hop query but involves a multi-hop relationship through \texttt{hasTimeseriesReference}. LLM-based KGQA tools often fail to resolve these intermediary relationships, requiring advanced methods for path discovery.

\textit{Property Extraction.} Some queries involve property-level connections, such as \texttt{value} or \texttt{hasUnits}. For instance, while AUTOKGQA identified a building’s area node in $Q_{id}=3,4$, it failed to retrieve its \texttt{brick:value}. KGQA tools must not only identify nodes but also explore their graph structure to resolve such property connections. Without this capability, query results remain incomplete, limiting real-world utility.

Overall, KGQA for CPS-IoT must offer rich language channels while demonstrating the ability to map vague concepts to diverse schema elements, balancing relevance and efficiency in subgraph extraction, resolving complex multi-hop relationships without explicit input definitions, extracting specific property-level details from implicit references, and enabling generalization to multiple uses or tasks, all while maintaining trust and ensuring sensor/language alignment.

%% file: sections/4-QuoVadis.tex
\section{Quo vadis CPS-IoT FMs?}
\label{sec:quo_vadis}
\noindent
Despite tremendous advancements in FMs and LLMs for CPS-IoT, the unique characteristics of the domain pose significant limitations.
We identify two distinct sets of challenges:
(a) technical challenges limiting FMs efficacy for CPS-IoT applications, highlighted as recommendations upon the experiments presented in Section~\ref{sec:our-insights};
and (b) challenges in developing an ecosystem of software tools, datasets, and benchmarks that facilitate rapid ideate-develop-validate iterations.  



\subsection{Desiderata for FMs \& LLMs in CPS-IoT}

\subsubsection{Generalizability across sensor configurations.}~ 
Language and image modalities exhibit significant consistency and standardization in data representation and interpretation (e.g., words in a language, common image and video formats, etc.).
In contrast, sensor data varies significantly within and across modalities, lacking standardized representations and consistent information content. 
Viewing sensor time series as sequences of values overlooks the intricate dependence of information content on sensor configuration and physical channel.
Supervised discriminative models and recently emerging FMs for CPS-IoT enforce strict sensor configuration during training and anticipate similar configurations at test time.
Multimodal FMs encounter challenges with novel sensor positions and orientations, particularly for 3D spatial intelligence tasks~\cite{wu2024flexloc}.
CPS-IoT FMs must ingest diverse sensor data and generalize to unseen sensor configurations. This requires architecture and training frameworks that learn sensor representations aware of configurations,  either inferred from data or provided via appropriate metadata channels.

\subsubsection{Generalizability across CPS-IoT tasks.}
~
Existing CPS-IoT FMs also fail to generalize across the broad spectrum of tasks encountered in CPS-IoT.
As noted in Section~\ref{sec:sota}, current CPS-IoT FMs -- both end-to-end models performing entire tasks (e.g.,~\cite{ansari2024chronos, liu2024unitime, goswami2024moment}) and those mapping sensor data to embeddedings for use with task-specific heads (e.g.,~\cite{xu2021limu, liu2023focal, kara2024freqmae, narayanswamy2024scaling}) --  primarily focus on a single or a small set of tasks, such as forecasting, imputation, and classification.
Other than the attempts to use LLMs as general-purpose CPS-IoT FMs, which have been unsatisfactory in both performance and resource usage, no CPS-IoT FMs have yet emerged that can handle the myriad of CPS-IoT tasks, such as complex spatiotemporal events, planning, reasoning, control, etc.
The main reason is that current CPS-IoT FMs pursue the low-hanging fruit of adapting advanced language and vision models to sensor data that could be treated as sequences, whereas FMs that generalize to a broader array of CPS-IoT tasks will require more general capabilities of reasoning, optimization, and planning in 4D space-time.


\subsubsection{Sensor Tokenization.}~ 
CPS-IoT FMs struggle to segment continuous input modalities effectively. 
They often require tokenizing and segmenting long-range sensor data while preserving semantic meaning and local granularities, which is challenging when adapting existing AI systems originally designed for language and vision modalities that have clear tokenization.
For example, in NLP, tokenization occurs at the word or sub-word level~\citep{radford2021learning}, and in computer vision, it is applied at the level of individual objects within an image~\citep{dosovitskiy2020image}.
Existing models (e.g., transformers~\citep{vaswani2017attention}) are designed for discrete tokens with natural vocabulary and segmentation and require preprocessing to segment and discretize continuous sensing signals, which often leads to information loss~\citep{spathis2024first}.
Earlier work in statistics and signal processing on change point detection~\cite{aminikhanghahi2017survey,tartakovsky2012efficient} can model the semantic changes in information, such as a sudden change in physical activity frequency, structural collapses, or video scene transitions.
Extending these methods for long and continuous sensor data and integrating them end-to-end with CPS-IoT FMs remains challenging.
Key questions include whether CPS-IoT FMs should shift focus from sequences of samples representing observed states to key events (state changes) and learn to represent their spatiotemporal relationships.

\subsubsection{Models for continuous and long sensor streams.} 
~Another challenge for CPS-IoT FMs is learning representations from long, continuous sensor streams, particularly for tasks like complex event detection that require reasoning over long periods (~\cite{xing2020neuroplex, ren2021synergy}.
Transformers, while pervasive, scale poorly with respect to sequence length~\cite{ben2024decimamba}.
As a result, many long-range~\citep{beltagy2020longformer}, memory-augmented~\citep{bulatov2022recurrent}, and sparse attention~\citep{child2019generating} transformers have been proposed to process long-range sensor data.
State space models~\cite{gu2023mamba, zhu2024vision} are another recently proposed alternative to transformer-based discrete sequence models by leveraging continuous dynamic systems (i.e., states), enabling them to scale effectively to very long sequence lengths and continuous data.
A related challenge is learning information from multiple streams of long-range sensor data, where the task-specific information varies over time~\cite{liang2023quantifying}.
Future work should extend these long-context models (e.g., long-range transformers, memory-based models, state space models) to better capture relationships and interactions between multiple sensor streams that may not be aligned in time or space~\cite{liang2022high,liang2024foundations,tsai2019multimodal}.


\subsubsection{Ability to incorporate human knowledge: Neurosymbolic Architectures.}~
Current AI systems for language and vision interact only with humans, while CPS-IoT systems also engage with engineered systems with well-understood physics, albeit with perturbations.
CPS-IoT systems are also designed to meet precise safety and other requirements, unlike LLM-based chatbots with ambiguous and subjective guardrails.
CPS-IoT FMs can leverage human knowledge about system physics and operating constraints to simplify development, especially in data-sparse settings, and enable rapid adaptation to deployments and changing requirements.
This is particularly useful in settings with deployment-to-deployment variation and short-lived deployments. 
While LLM can process human knowledge via prompts at runtime, they lack sufficient reliability and struggle with complex logical reasoning.
An alternative is neurosymbolic~\cite{garcez2019neural} CPS-IoT FMs, which combine neural layers with symbolic layers that encode human knowledge and algorithms while capable of run-time fine-tuning.
While neurosymbolic models have shown promise in simple CPS-IoT settings~\cite{xing2020neuroplex}, considerable challenges remain: computational scaling of symbolic stages; efficient management of uncertainty, noise, and spatiotemporal concepts;  methods for efficient training and fine-tuning~\cite{cunnington2024role}; and extending beyond perception~\cite{mitchener2022detect}.

\subsubsection{Rich language channel.}~
Many of these desiderata depend on the CPS-IoT FM having access to auxiliary information at runtime beyond sensory data, such as task description, static context (e.g., sensor configurations), and dynamic context (e.g., environment conditions).
Instead of dedicated typed channels for each auxiliary information entity, CPS-IoT FMs could benefit from a general-purpose language channel that integrates tokens from auxiliary data with the sensor data stream when mapping to the latent embedding space.
Aligning the language channel with sensor data unifies reasoning and prediction over sensor data and leads to more versatile, explainable, and adaptable models
Preliminary evidence~\cite{chattopadhyay2024context} demonstrates the value of such contextual information in time series forecasting tasks.
Collecting rich annotations of various sensor inputs through language can enable the development of a unified language to describe, reason, and predict on these sensors~\citep{mo2024iot}, just like how unified language descriptions of visual images, objects, and relationships have catalyzed vision language models~\citep{gan2022vision,liang2024foundations}.

\subsubsection{New architectural abstractions.}~
For CPS-IoT models to gain wide adoption and serve as reusable building blocks for system designers, they must be easy to integrate and optimize with system software and hardware and effectively interact with other custom FMs.
Moreover, as we discussed in Section~\ref{sec:resource_quality_tradeoffs}, FMs, due to their generality, require significant resources in absolute terms and compared to specialized single-purpose models.
Therefore, the current practice of each application in a multi-tenant CPS-IoT system bundling its own models is unsustainable with FMs, particularly as most CPS-IoT systems cannot rely on cloud services for various reasons.
Therefore, integrating CPS-IoT FMs into the overall system stack requires careful consideration.
Given their resource footprint, we believe CPS-IoT FMs will be run-time composable services provided by the platform (i.e., hardware + OS) and shared by multiple applications rather than merely design-time reusable blocks. 
This ecosystem requires standardized yet flexible interfaces for CPS-IoT-FMs-as-system-service, along with autotunable models in accuracy-resource-performance space based on the needs of subscribing applications.
An example of such CPS-IoT models is the M4 composable multimodal FM~\cite{yuan2024mobile}.

\subsection{CPS-IoT FM Development Ecosystem} 
\noindent
Developing quality FMs is difficult. Yet, in language and vision domains, the co-development of open models, large datasets, diverse benchmarks, and supporting tools has accelerated development.
For CPS-IoT FMs with the aforementioned desiderata to emerge, a similar ecosystem is needed to facilitate a virtuous cycle of ideate-develop-validate iterations and application-level development.
In particular, there is a need for large-scale unified benchmarks spanning multiple CPS-IoT domains, sensing modalities and tasks.
Existing CPS-IoT datasets~\citep{Ahuja2021TouchPose, geiger2013kitti, kong2021EyeMU, Mollyn2022samosa, Riku2022RGBDGaze}, are often constrained into a few of these dimensions.

One challenge in creating an organic collaboration for CPS-IoT is the non-human-interpretable nature of its data, particularly raw sensory input.
Large vision and language models are powered by the abundant internet-scale text, image, and video data aligned with human perception and easily shared online, creating natural incentives for publishing raw data. 
In contrast, most CPS-IoT sensory data lacks these incentives.
Although highly processed, human-interpretable data (e.g., fitness statistics) can be shared, raw sensory data (e.g., IMU) is seldom disclosed.
Privacy, intellectual property concerns, sensor heterogeneity, and data volume exacerbate the issue.
Most importantly, non-interpretable data offers limited immediate value, significantly reducing incentives for sharing.
To create high-quality CPS-IoT FMs, we must incentivize individuals and organizations to share raw sensor and actuator data by offering socially constructed derivative information or new ways for people to experience the raw data that align with their senses.   

A second factor is that CPS-IoT systems vary widely in task, sensing, platform, actuator, environment, and safety requirements owing to their diverse applications (e.g., agriculture, aeronautics, buildings, and healthcare). Although all of these systems perform PCCA loops, a single CPS-IoT FM to handle all of them is impractical. However, drawing inspiration from robotics success, targeting application-specific CPS-IoT FMs for energy, healthcare, and transportation may be feasible.
These \textit{micro foundation models} ($\mu$FMs) are more resource efficient and can be embedded.
Although the diversity of each domain still surpasses that of robotics, the development of $\mu$FM is more manageable.

%% file: sections/5-Conclusions.tex
\section{Conclusions}
\noindent
We explored the potential of foundation models (FMs) and large language models (LLMs) in enhancing Cyber-Physical Systems (CPS) and the Internet of Things (IoT) by addressing challenges in perception, cognition, communication, and action (PCCA) loops. While traditional task-specific machine learning models in CPS-IoT have limitations due to data annotation requirements, sensor heterogeneity, and deployment constraints, FMs offer a transformative approach with their task-agnostic adaptability and self-supervised learning methods. However, while FMs and LLMs hold immense promise for CPS-IoT, their practical adoption require going beyond current approaches that simplistically adapt methods from general NLP and computer vision domains to CPS-IoT. Instead, CPS-IoT FMs, and LLMs targeting CPS-IoT applications, must account for the unique characteristics of sensory data, physical actions, and tasks across the PCCA loops. Not only are the research problems challenging, but solving them would also require community-scale effort at creating a unified ecosystem of data sets, models, and benchmarks. By addressing these gaps, FMs and LLMs could serve as foundational tools for the next generation of CPS-IoT applications, fostering widespread innovation across diverse domains.